\pdfoutput=1
\documentclass[letterpaper, 10 pt, conference]{ieeeconf}
\IEEEoverridecommandlockouts

\IEEEoverridecommandlockouts

\usepackage[utf8]{inputenc} 
\usepackage[T1]{fontenc}    
\usepackage{url}            
\usepackage{nicefrac}       
\usepackage{microtype}      
\usepackage[english]{babel}

\usepackage{graphicx}
\usepackage{amsmath}
\usepackage{amssymb}
\usepackage{booktabs}
\usepackage{mathtools}
\usepackage{amsmath,amssymb,amsfonts}
\usepackage{array}
\usepackage{algorithmic}
\usepackage{graphicx}
\usepackage{textcomp}
\usepackage{multirow}
\usepackage{adjustbox}
\usepackage[font=scriptsize]{caption}
\usepackage{subcaption}
\usepackage{xcolor}
\usepackage{bm}
\usepackage{soul}
\usepackage{booktabs}
\usepackage{tikz} 
\usepackage[colorlinks=true, allcolors=blue]{hyperref}
\usepackage{makecell}
\usepackage{duckuments}
\usepackage{comment}
\usepackage[normalem]{ulem}
\usepackage{comment}
\usepackage{xcolor}
\usepackage[margin=0.75in]{geometry}

\usepackage{hyperref}       

\begin{document}

\title{NeRFifyMesh: Optimizing Neural Radiance Fields from Textured Meshes for Robotics Scene Building}

\author{Nillan  Nimal$^{1}$, Mahboubeh Asadi$^{2}$, and Sajad Saeedi$^{3}$ 
\thanks{$^{1}$ Department of Mechanical and Industrial Engineering, University of Toronto, Canada}
\thanks{$^{2}$Department of Mechanical, Industrial and Mechatronics Engineering, Toronto Metropolitan University, Canada}
\thanks{$^{3}$Department of Computer Science, University College London, UK}
}


\maketitle

\begin{abstract}
In robotics, scene representation plays a pivotal role
in understanding and interacting with the environment. The
advent of Neural Radiance Fields (NeRF) and its variants, as
a novel representation, has opened a new frontier of research.
In applications such as semantic mapping and simulation, roboticists aim to build scenes using multiple NeRF models, each representing an object. While extensive datasets of 3D mesh models already exist, there is an urgent need to develop tools to convert these assets to NeRF models for rapid algorithm development and testing. This paper presents a new pipeline for converting existing mesh models to NeRF representations by artificially generating a ground truth point-based radiance field through sampling mesh geometry and texture. This approach alleviates the need for camera-based sampling or rendering multi-view images of the original mesh to train the NeRF model. Extensive benchmarking demonstrates that our method yields comparable rendering quality to the baselines. Additionally, the application of this representation is shown by constructing unified NeRF scenes and performing collision simulations with extracted geometry. 

\end{abstract}

\maketitle

\section{Introduction}

In 3D digital graphics and computer vision, the accurate and efficient representation of objects and scenes is paramount. Popular methods for representing 3D structures vary based on their characteristics and applications. Point clouds capture geometric detail sparsely through individual data points. Meshes define surface geometry by using interconnected points, or vertices, which together form polygonal units like triangles~\cite{feng2018meshnet}. Voxels offer a distinct approach by discretizing space into uniform, grid-like structures \cite{voxel}. Signed Distance Functions (SDFs) represent the distance of a point from the surface of a 3D shape where the sign is negative inside the surface and positive outside \cite{song2010hierarchical}. Recently, Neural Radiance Fields (NeRFs) have emerged as a novel representation~\cite{mildenhall2020nerf}. Optimized for visual consistency, NeRFs can achieve state-of-the-art results for novel view synthesis and image-based rendering tasks. 
 
The ability to transition between diverse model representations can be immensely valuable. Different fields and applications often require or benefit from specific 3D representations. For instance, while robotics might leverage SDFs for object interaction, computer graphics might prefer mesh models for detailed visual rendering. As such, the interoperability between these representations can bridge the gap between fields and facilitate a wider range of applications. Converting between mesh and NeRF format is particularly interesting given the prevalence of mesh models in computer graphics and robotics. Textured mesh models are widely available in databases such as Amazon Berkeley Objects \cite{collins2022abo}, Google Scanned Objects \cite{downs2022google}, ShapeNet \cite{chang2015shapenet}, and Objaverse~\cite{objaverse}.

{Developing an efficient and automated representation conversion tool has many applications in robotic systems, such as composing large-scale scenes out of many NeRF models, where the NeRF models are built from existing mesh models.} However, traditional pipelines to convert representations entail rendering images of the mesh model from known camera poses that encompass the entire model to serve as the ground truth for optimizing a NeRF. This approach has significant drawbacks. Rendering high-quality images from mesh models is computationally expensive and training images that omit areas of the model can lead to poor results. {Additionally, the rendered images require storage, which can be problematic when working with large-scale datasets such as Objaverse~\cite{objaverse}.}

\begin{figure}
\centering
\includegraphics[width=\linewidth]{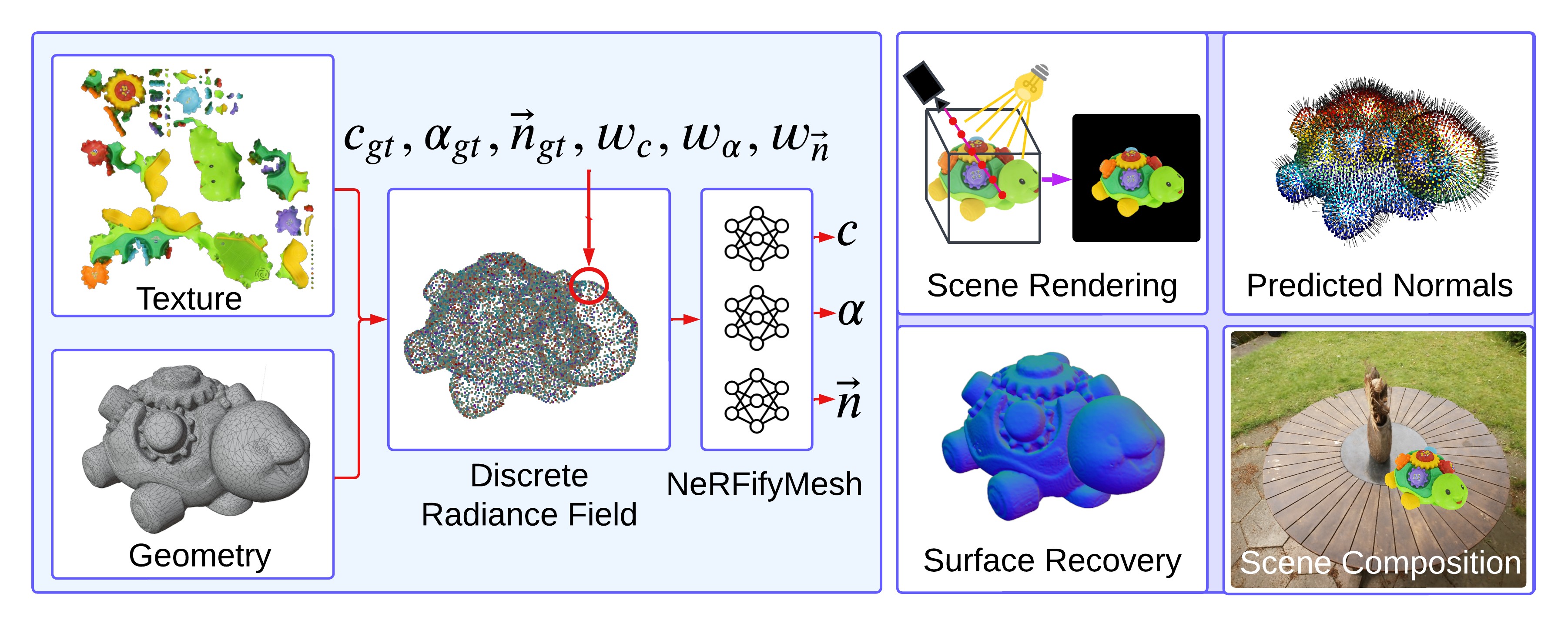} %
\vspace{-6mm}
\caption{We present NeRFifyMesh, an architecture and pipeline for converting textured meshes into Neural Radiance Fields (NeRFs). We generate a ground-truth radiance field from the mesh, encompassing color, opacity, point normals, and their respective weighting terms. 
NeRFifyMesh supports scene rendering, normal prediction, surface recovery, and scene composition.
\vspace{-5mm}
}
\label{fig:mainfigure} 
\end{figure}


In this paper, we propose NeRFifyMesh, a pipeline that removes the rendering optimization step from the training process and alleviates the need for high-quality ground-truth images and camera poses of the mesh model (see Fig.~\ref{fig:mainfigure}). We achieve this by sampling the mesh to generate a discrete point-based radiance field with the properties of color ($\textbf{c}_{\text{gt}}$), binary opacity ($\alpha_{\text{gt}}$), point normals (\(\vec{n}_{\text{gt}}\)), and their significance, $w_c $, $w_\alpha$, and $w_{\vec{n}}$ respectively. We then proceed to train our neural representation directly on this discrete field. By including point normals in our representation, we can relight our rendered scene post-training, recover accurate geometry through surface reconstruction, and facilitate applications that rely on geometric information, such as robotics simulation. The contributions of this work are:
    1) A mesh-derived Neural Radiance Field representation is proposed, where neural supervision is performed on a discrete radiance field generated from mesh geometry and texture. This method alleviates the need for multi-view image supervision or camera-pose information during training.    
    2) Surface normals are learned by the representation, enabling normal-based shading during rendering and supporting geometry-dependent applications. 
    3) The effectiveness of the representation is demonstrated through model compression, relighting, surface extraction and collision simulation experiments in constructed scenes.  
The project page and videos can be found at: \href{https://sites.google.com/view/nerfifymesh}{https://sites.google.com/view/nerfifymesh}


{The rest of the paper is organized as follows.
Sec.~\ref{sec:related} presents the related works. 
Sec.~\ref{sec:method} 
describes the proposed method.
Sec.~\ref{sec:exp} presents the results.
Sec.~\ref{sec:con} concludes the paper.
}

\

\section{Related Work}\label{sec:related}
This section presents the literature review 
for neural representations, mesh-based NeRF pipelines, and NeRFs in robotics applications.  

\subsection{Neural Implicit Representations}
{Neural Radiance Fields (NeRFs)~\cite{mildenhall2020nerf} are a novel representation, 
mapping a 2D ray viewing direction ($\phi, \theta)$ and Cartesian coordinates for a 3D location ($x, y, z$) to volume density ($\sigma$) and emitted view-dependent radiance ($r,g,b$). 
Other neural implicit representations can also parameterize the geometry and/or appearance of meshes in a continuous and compact manner. Neural SDFs~\cite{park2019deepsdf,Siren} model surfaces implicitly as the zero-level set of signed distance functions. For example, Neural Geometric Level of Detail (NGLoD) \cite{nglod} represents a neural SDF as a collection of feature vectors in a sparse voxel octree (SVO) where the levels of the SVO define levels of detail. 
Occupancy Networks \cite{mescheder2019occupancy} employs a neural network-based binary classifier to implicitly define surfaces as a set of 3D points. Instant-NGP \cite{instantngp} addresses the computational costs of utilizing fully connected neural networks by introducing a multi-resolution hash encoding that allows for accelerated training and rendering of neural graphics primitives such as NeRF and Neural SDFs.

%

In addition, mesh autoencoders 
\cite{autoencode1,autoencoder2} apply convolutional operations to meshes directly to create latent representations for mesh reconstruction. Although these works either directly utilize mesh data or derive point clouds from sampled meshes, their focus is limited to mesh geometry encoding and reconstruction. 


Volume rendering-based methods \cite{wang2023neus,yu2021plenoctrees,nsvf} and Scene Representation Networks (SRNs) \cite{scenerepresentation} take a different approach. They represent 3D scene geometry and appearance derived from multi-view images. NeuS presents a volume rendering method for surface reconstruction from multi-view images that addresses the limitations of NeRF which lacks surface optimization \cite{wang2023neus}. NeuMesh~\cite{yang2022neumesh}, which is supervised with NeuS, utilizes a mesh-based scaffold with learnable disentangled geometry and texture features to encode a neural implicit field. 
These approaches require the synthetic generation of images from the mesh model. This leads to reconstruction inaccuracies for regions of the mesh that are rarely visible or occluded in the generated images.

\subsection{Mesh-based NeRF Pipelines}

The extraction of accurate underlying 3D geometry from NeRFs presents a challenging problem given that they are not explicitly optimized for geometry. Consequently, current works \cite{nerfmeshing,delicatetextured}, have developed methods for extracting meshes from volumetric NeRF representations, enabling integration with common computer graphics pipelines and real-time rendering. NeRFMeshing \cite{nerfmeshing} utilizes a signed surface approximation network (SSAN) to represent a triangle mesh of the scene and an appearance network to capture view-dependent colours. NeRF2Mesh \cite{delicatetextured} extracts meshes with view-independent diffuse textures and view-dependent specular textures. 
Additional approaches incorporate a mesh representation of the target scene to edit NeRFs \cite{yuan2022nerfediting,denerf,NeRFshop} or improve rendering efficiency \cite{chen2023mobilenerf}. For example, DE-NeRF \cite{denerf} attaches learnable disentangled features to an extracted triangle mesh of the scene to allow for editing the model.



Mesh2NeRF~\cite{chen2024mesh2nerf} showed that training directly on meshes significantly
improves rendering quality. However, their method samples from virtual cameras
rather than mesh geometry, does not learn surface normals, bakes in
lighting, and relies on off-the-shelf NeRF backbones. Our representation
instead uses a custom architecture that leverages mesh geometry directly,
enabling relighting and accurate geometry extraction through learned
surface normals.


\subsection{NeRFs in Robotics Applications}

There is an increasing prevalence of NeRFs in robotics due to their continuous scene representation and photorealistic rendering capabilities. For robotics applications, NeRFs have been leveraged for data generation, perception, mapping, simulation, and multi-robot agents~\cite{Di-NeRF}.

Neural Sim \cite{ge2022neural_SIM}, presents a pipeline to generate synthetic data with NeRFs to train vision models, enabling applications such as object detection and scene understanding. Similarly, Wang \textit{et al.} \cite{SemanticNeRF} investigates the performance of a NeRF model that uses only semantic data for tasks such as segmentation and detection.

Additionally, NeRFs have demonstrated potential in robot navigation and mapping. Adamkiewicz \textit{et al.} \cite{Nerf_vision_Robot_sim} demonstrated vision-only navigation of robots in a learned radiance field, with trajectory algorithms to avoid high-density regions corresponding to obstacles.


In terms of physics-based simulations, Cleac'h \textit{et al.} \cite{Differentiable_Physics_Simulation} produce a differentiable neural object pipeline which includes NeRFs for simulating the motion of objects. This allows for the estimation of dynamics properties such as friction and mass from visual data. NeRF2Real \cite{nerf2real} synthesized scenes composed from static NeRF environments and dynamic objects and utilized a physics-engine to compute contact dynamics for sim2real policy transfer.


Together these works demonstrate the advancing interest of NeRFs in robotics. Consequently, a pipeline to convert meshes to NeRFs will be useful by allowing existing mesh assets to be integrated into NeRF-based simulation and perception frameworks.

\section{Method}\label{sec:method}
NeRFifyMesh aims to leverage the explicit geometry and texture information of a mesh model to encode a mesh within a NeRF representation without the need for multi-view images. 
Here, Sec.~\ref{sec:representation} provides an overview. 
Network architecture and loss functions are covered in 
Sec.~\ref{sec:architecture} and Sec.~\ref{sec:loss}, respectively. 
Rendering is explained in Sec.~\ref{sec:rendering}. 

\subsection{{Overview:} NeRFifyMesh Representation}
\label{sec:representation}
The NeRFifyMesh representation makes two assumptions: 
1) mesh appearance is view-independent, 2) meshes are opaque with no translucent features. 
\noindent Consequently, 2D ray view-direction parameters ($\phi, \theta)$ are omitted, and opacity is modeled as a binary function, as in Occupancy Networks \cite{mescheder2019occupancy}. Since normals are learned in this representation, view-dependent shading can be incorporated during rendering (Sec.~\ref{sec:relighting}). 

NeRFifyMesh artificially generates a point-based discrete radiance field by sampling points within the bounding box of the mesh model $\Omega$ and calculating the signed distance. The color, opacity, and point normal assignment for each sampled point \( (x,y,z) \in \Omega \) is derived from the SDF values: 
\begin{equation}
\textbf{c}_{\text{gt}} = 
\begin{cases} 
\textbf{c}_S (x', y', z') \\
\textbf{c}_S (x, y, z)  \\
0 \\
\end{cases},~
\vec{n}_{\text{gt}} = 
\begin{cases} 
\vec{n}_S (x', y', z') & \text{\scriptsize SDF} < 0 \\
\vec{n}_S (x, y, z) & \text{\scriptsize SDF} = 0 \\
0 & \text{\scriptsize SDF} > 0\\
\end{cases},~
\end{equation}
\begin{equation}
\alpha_{\text{gt}}(x,y,z) = 
\begin{cases} 
1 & \text{\scriptsize SDF} \leq 0 \\
0 & \text{\scriptsize SDF} > 0\\
\end{cases}
\nonumber
\end{equation}

\noindent where $\textbf{c}_S$ and $\vec{n}_S$ is the color and normal of the mesh surface and $(x',y',z')$ is the nearest point on $S$. Surface colors are obtained from the associated texture image and surface normals are obtained through barycentric interpolation from the vertex normals. Defining colors for points within the mesh is essential for accurate volume rendering since samples along the camera rays might bypass the mesh's implicit surface. As a result, internal mesh points could influence the final rendered color, leading to unpredictable rendering behavior. 

\noindent \subsubsection{Importance Weighting} A direct training approach on the extracted colors, normals, and volume densities of this discrete radiance field would prove suboptimal. This is because the importance levels associated with the points vary. For example,  the probability of a ray terminating increases proportionally with the distance inside the mesh, and thus, the accuracy of color, opacity, and normal values of interior points is less critical as the distance from the mesh surface increases. Moreover, since external mesh points inherently possess zero opacity (empty space) their colors/normals should have zero contribution to the final rendering. Therefore, inspired by Davies \textit{et al.} \cite{davies2021effectiveness}, we incorporated weighting terms to modulate the influence of the color, normals, and opacity values in our representation. The weights are also derived from the computed SDF values of the sampled points:
\begin{equation}
w_c(x,y,z) = w_{\vec{n}} (x,y,z) = 
\begin{cases} 
\exp\left(\frac{-|SDF|}{\tau}\right)& \text{\scriptsize if SDF} \leq 0 \\
0 & \text{\scriptsize if SDF} > 0\\
\end{cases}~,
\label{colorweights}
\end{equation}
\begin{equation}
w_\alpha(x,y,z) = 
\begin{cases} 
\exp\left(\frac{-|SDF|}{\tau}\right)& \text{\scriptsize if SDF} \leq 0 \\
1 & \text{\scriptsize if SDF} > 0\\
\end{cases}~
\label{opacityweights}
\end{equation}

 \noindent The threshold parameter $\tau$ controls the rate of decay of the weights as the SDF values move away from 0 (the surface) within the mesh. We set $\tau= 0.1$. This weighting scheme ensures that interior colors and opacities that are closer to the mesh surface are prioritized. 
 



\begin{figure}
\centering
\includegraphics[width=\linewidth]{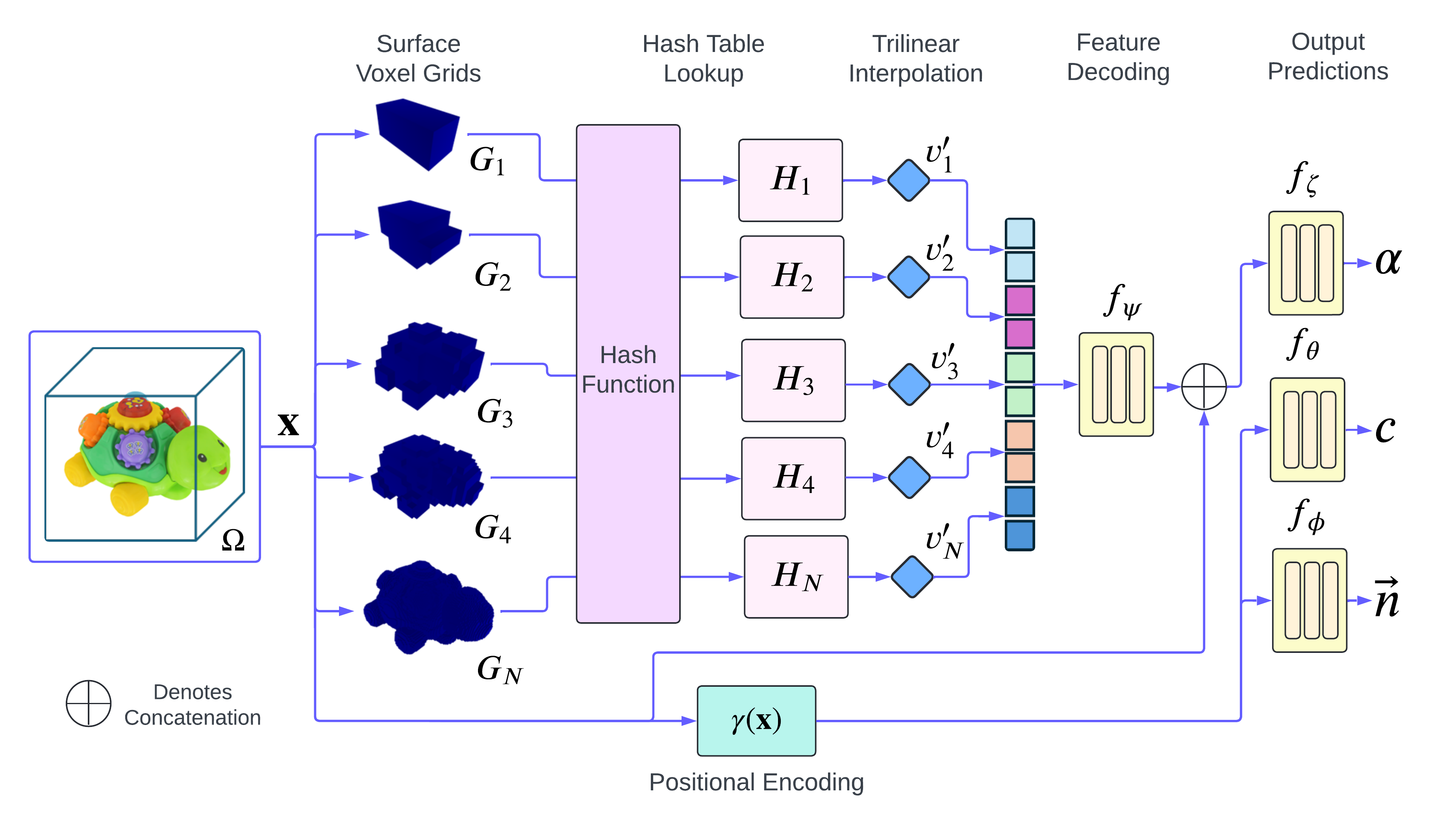} %
 \caption{An overview of the NeRFifyMesh network architecture. Input points \(\textbf{x}\)  are queried across multi-resolution surface voxel grids with hash-encoded features. Interpolated features are decoded with network \(f_{\psi}\) and combined with the original input to predict opacity through network \(f_{\zeta}\). Positional encoding of the input is provided to networks \(f_{\theta}\) and \(f_{\phi}\) to predict colors and normals, respectively.}


\label{fig:network} 
\end{figure}
\subsection{Network Architecture}
\label{sec:architecture}

The NeRFifyMesh architecture decouples predicted color \(\textbf{c}\), point normals \(\vec{n}\), and opacity \(\alpha\) into three separate networks. Each network is optimized with separate loss functions and trained simultaneously for efficiency. An overview of the network architecture is displayed in Fig.~\ref{fig:network}.


\subsubsection{Color and Point Normal Networks}
For color and normals prediction, positional encoding transforms the input point \(\textbf{x}\) into a higher-dimensional space using a series of sine and cosine functions \cite{mildenhall2020nerf}. 
%
\noindent Positional encoding, as implemented by Mildenhall \textit{et al.} in the original NeRF \cite{mildenhall2020nerf}, allows for the representation of high-frequency details by capturing the fine spatial variations in the scene. The positionally encoded input is then fed into separate color and normal decoder networks, \(f_\theta\) and \(f_\phi\) respectively. Each network is composed of SIREN (Sinusoidal Representation Networks) layers. SIREN layers use sinusoidal activation functions which allows for the effective modelling of complex continuous signals, including those with high frequency components \cite{Siren}. This structure captures color and normal variations across the mesh.


\subsubsection{Opacity Network}

For opacity prediction, we introduce a multi-resolution hash-encoding architecture that directly leverages mesh data. The surface points sampled from the mesh are utilized to initialize \(N\) surface voxel grids \(G_1, G_2, ..., G_N\). Each grid \(G_l\) is set to a resolution of \(R=2^l\) where \(l\) is the level of detail. Trainable latent vectors \(v_l[i,j,k]\) with 2 dimensions are assigned to each grid vertex where \(i,j,k\) are the vertex indices. For a given input point \(\textbf{x}\),  the point is queried at each of the surface voxel grids by multiplying by the grid resolution. If \(\textbf{x} \in G_l\), the indices of the eight corners that the point \(\textbf{x}\) belongs to are used as hash keys to retrieve the corresponding latent vectors from hash table \(H_l\),  i.e., \(v_l[i,j,k] = H_l([i,j,k])\). Trilinear interpolation is utilized to interpolate the latent vector according to its position in the voxel grid producing \(v'_l(\textbf{x})\). If the point \(\textbf{x} \not\in G_i\), then \(v'_l(\textbf{x}) = v_{l, default}\) where \(v_{l, default}\) is a predefined default latent for \(G_l\). These latent vectors for \(N\) grid levels are concatenated together, \(V(\textbf{x}) = [v'_1(\textbf{x}), v'_2(\textbf{x}), ...,v'_N(\textbf{x})]\) and fed as input to the feature decoder network \(f_\psi\). Lastly, the output from the feature decoder network is concatenated together with \(\textbf{x}\) and passed to the density network \(f_\zeta\) to predict the opacity \(\alpha(\textbf{x}) = f_\zeta([f_\psi(V(\textbf{x})), \textbf{x}])\).

The hash-encoding architecture employed for NeRFifyMesh is inspired by Instant-NGP \cite{instantngp}. However, as opposed to initializing trainable latent vectors for each grid point at a given resolution, NeRFifyMesh strategically assigns these vectors solely to the grid points corresponding to the occupied voxels within a surface voxel grid. Points external to this grid are mapped to a default latent vector, effectively distinguishing the mesh-contained regions from the surrounding empty space. This approach not only leverages the inherent geometric information provided by the mesh to minimize the number of trainable parameters but also integrates this structural data into the implicit representation. The default latent vector plays a crucial role in assisting the decoder network to differentiate between the mesh surface and void regions effectively. 

Since the surface voxel grid is initialized from the mesh, to avoid the necessity of retaining the original mesh to rebuild the model, the indices \(I = [i,j,k]\) of the surface voxels at each grid \(G_l\) are saved with the model. 


\subsection{Loss Functions}
\label{sec:loss}
The color network is optimized via a weighted mean-squared error (MSE) loss between the predicted and ground-truth colors ($\mathbf{c}$ and $\mathbf{c}_{\text{gt}}$), using the color weights defined in Eq.~\eqref{colorweights}:
\begin{equation}
\mathcal{L}_{c} = \frac{1}{n} \sum_{i=1}^n w_{c_i} ({\textbf{c}}_i - \textbf{c}_{\text{gt}, i})^2
\end{equation}
\noindent The normal network is supervised using a weighted mean squared cosine similarity function, with the weights of Eq.~\eqref{colorweights}:
\begin{equation}
\mathcal{L}_{\vec{n}} = \frac{1}{n} \sum_{i=1}^{n} \left( w_{\vec{n_i}} (1 - \cos{\delta_i})^2 \right),~~
\cos{\delta_i} = \vec{n}_{\text{gt}, i} \cdot \vec{n}_i
\end{equation}


The optimization of the opacity network and the trainable latent vectors is conducted through a composite loss function. This function comprises two primary components. The first component is a weighted binary cross entropy loss (BCE) calculated from predicted and ground-truth opacities (\(\alpha\) and \(\alpha_{\text{gt}}\)) that uses opacity weights \(w_{\alpha}\) from Eq.~\eqref{opacityweights}:


\begin{equation}
\mathcal{L}_{\text{BCE}} = -\frac{1}{n} \sum_{i=1}^n w_{\alpha_{i}} [ \alpha_{\text{gt},i} \log\alpha_i + (1 - \alpha_{\text{gt},i}) \log(1 - \alpha_i) ]
\end{equation}

\noindent The second component is a regularization term that aggregates the norms of the trainable latent vectors across all grid levels~\(N\):
\begin{equation}
\mathcal{L}_{\text{reg}} = \sum_{l=1}^N \| v_l \|^2
\end{equation}
\noindent The regularization term is designed to prevent overfitting by penalizing the magnitudes of the latent vector values. Consequently, the total opacity loss is expressed as the sum of the BCE and regularization losses:
$\mathcal{L}_{\alpha} =  \lambda_1\mathcal{L}_{\text{BCE}} + \lambda_2\mathcal{L}_{\text{reg}}$, 
\noindent where \(\lambda_1\) and \(\lambda_2\) are loss weighting terms.

\subsection{Rendering} \label{sec:rendering}

 The rendering process involves sampling a set of points along camera rays, evaluating the NeRFifyMesh model at these points, and then integrating the results to produce the final pixel colors. Similar to Mesh2NeRF~\cite{chen2024mesh2nerf}, our method directly predicts per-sample opacities \(\alpha\) rather than densities \(\sigma\). This choice reflects the fact that opacity can be simply derived from Signed Distance Fields (SDFs). 

Given sampled points \(\{t_i\}_{i=1}^M\) with interval distances \(\delta_i = t_{i+1} - t_i\), the accumulated transmittance is:
\begin{equation}
T_i = \prod_{j=1}^{i-1} (1 - \alpha_j)
\end{equation}
The final pixel color for a given camera ray $\textbf{r}$ is then approximated from predicted sample opacities and colors as:
\begin{equation}
\hat{\textbf{C}}(\textbf{r}) \approx \sum_{i=1}^M T_i \alpha_i \textbf{c}_i
\end{equation}

\section{Experiments}\label{sec:exp}
This section presents experimental results for NeRFifyMesh. Implementation details are first described, followed by comparisons of mesh scene renderings against NeRF-based methods trained from images. Additional applications and simulation experiments are then presented, followed by a discussion of limitations.


\subsection{Implementation Details} \label{sec:implementation}
    For each mesh model used in the experiments, 2 million samples were taken of points inside, outside and on the mesh surface using the strategy proposed by Takikawa \textit{et al.} \cite{nglod} to generate training data. NeRFifyMesh was trained to 300,000 iterations with a batch size of 4096 samples using a system with 50 GB of RAM and an NVIDIA P100 GPU. Training takes roughly 2 hours per model and the best weights for the color, opacity, and normal networks are retained. 

\subsection{Mesh Scene Rendering}\label{sec:meshscenerendering}
   
To evaluate the performance of NeRFifyMesh on mesh scene rendering, we selected 13 objects from the Google Scanned Objects (GSO) dataset \cite{downs2022google}, 7 objects from Poly Haven \cite{polyhavenPolyHaven}, and 5 objects from the Objaverse dataset \cite{objaverse}. The GSO dataset features scans of household objects, represented as simplified and watertight meshes. The Poly Haven dataset includes artist-created 3D mesh assets. The Objaverse dataset comprises scans of museum and nature artifacts with complex geometries and high-resolution textures. The Objaverse dataset is particularly challenging due to its high vertex counts and non-manifold geometry, which complicates sampling.

 NeRFifyMesh's rendering performance is benchmarked against two NeRF methods: the original NeRF \cite{mildenhall2020nerf}, and Instant-NGP \cite{instantngp}. While these methods require a set of input images for training, NeRFifyMesh is trained directly from mesh data. For comparison, we rendered 100 training images and 20 test images at 1024$\times$1024 resolution of each mesh from camera poses following a spherical coordinate system. These rendered images were used to train the NeRF-based methods.
    
The quality of our renders for the test images is compared to those generated by the other methods in terms of PSNR, SSIM, and LPIPS. In Table \ref{tab:comparison}, our method achieves the lowest (best) LPIPS score for all three datasets, indicating superior perceptual similarity. In terms of the SSIM measure, our method demonstrates competitive performance, with higher scores than NeRF but slightly lower scores than Instant-NGP. For the PSNR metric, ours is lower (worse) than Instant-NGP and NeRF for all three datasets. This highlights a methodological difference. Instant-NGP and NeRF are optimized for photometric rendering loss from multi-view images. In contrast, our representation is trained from mesh-derived samples without volume rendering optimization during training. As a result, while our method does not explicitly minimize pixel-level photometric error, it preserves perceptually meaningful geometric and textural details shown by our strong LPIPS performance.

A qualitative comparison of mesh scene rendering is shown in Fig.~\ref{fig:comparison}. Our method represents fine mesh details with high fidelity resulting in visually superior renders compared to the baselines Instant-NGP and NeRF.

\begin{table}[ht]
    \centering
    \caption{Quantitative comparison of rendering performance metrics across different datasets using our method, Instant-NGP (I-NGP) \cite{instantngp}, and NeRF \cite{mildenhall2020nerf}.}
    \tabcolsep=0.05cm
    \begin{tabular}{lccccccccc}
        \toprule
        \text{Method} & \multicolumn{3}{c}{\text{GSO} \cite{downs2022google}} & \multicolumn{3}{c}{\text{Poly Haven \cite{polyhavenPolyHaven}}} & \multicolumn{3}{c}{\text{Objaverse \cite{objaverse}}} \\
        \cmidrule(lr){2-4} \cmidrule(lr){5-7} \cmidrule(lr){8-10}
        & \text{PSNR$\uparrow$} & \text{LPIPS$\downarrow$} & \text{SSIM$\uparrow$} 
        & \text{PSNR} & \text{LPIPS} & \text{SSIM}
        & \text{PSNR} & \text{LPIPS} & \text{SSIM} \\
        \midrule
        Ours         & 31.98 & \textbf{0.024} & 0.977 & 35.11 & \textbf{0.034} & 0.965 & 31.70 & \textbf{0.037} & 0.958 \\
        I-NGP  & \textbf{37.00} & 0.027 & \textbf{0.982} & \textbf{38.52} & 0.052 & \textbf{0.973} & \textbf{35.62} & 0.043 & \textbf{0.965} \\
        NeRF         & 34.12 & 0.054 & 0.965 & 36.86 & 0.083 & 0.956 & 32.20 & 0.083 & 0.927 \\
        \bottomrule
    \end{tabular}
    \label{tab:comparison}
\end{table}

\begin{figure}
\centering
\includegraphics[width=1.0\linewidth]{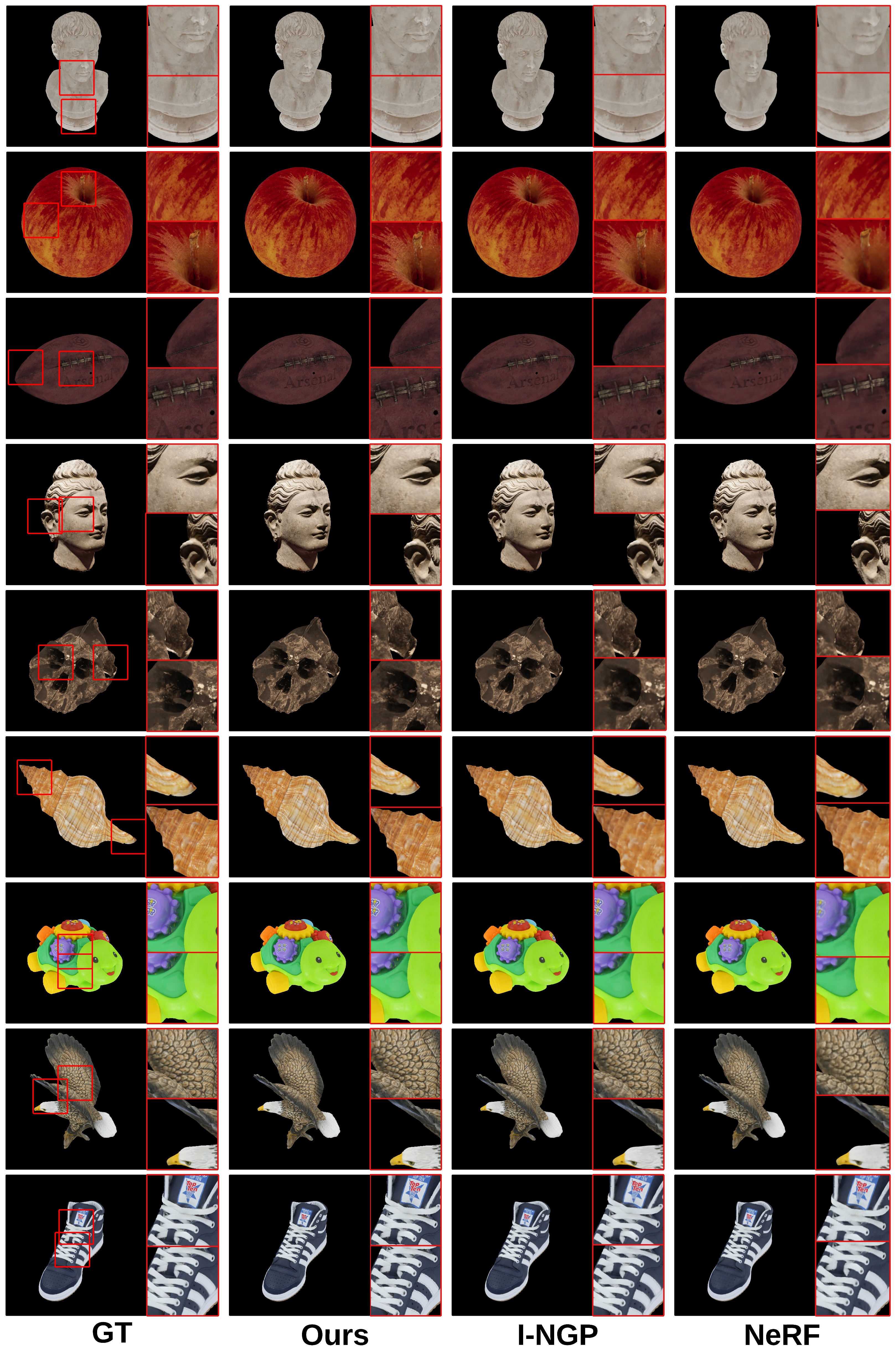} %
\caption{Visual comparison of mesh scene rendering results for our method, Instant-NGP \cite{instantngp}, and NeRF \cite{mildenhall2020nerf} across the GSO \cite{downs2022google}, Poly Haven \cite{polyhavenPolyHaven}, and Objaverse \cite{objaverse} datasets. Our method represents fine textural and geometric details with high fidelity.} 
\label{fig:comparison} 
\end{figure}

\subsection{Applications}\label{sec:applications}
\subsubsection{Mesh Compression}\label{sec:compression}
Given the size of many textured mesh datasets, size compression is a crucial application of neural mesh representations. To analyze model compression, the average total mesh size was measured (including geometry, texture, and material files) of the tested models for each of the three datasets in Sec.~\ref{sec:meshscenerendering}. These mesh sizes were compared to our representation, which includes the opacity, color, and normal module weights, the hash table keys, and the mesh bounding box used to reconstruct the model. 
Table \ref{tab:compression} shows that our representation achieves significant compression ratios compared to the baselines. The compact nature of our representation highlights the space efficiency of initializing surface voxel grids at different levels of detail using the original mesh. %

\begin{table}[ht]
    \centering
    \caption{Comparison of average model sizes and compression ratios across datasets.}
    \tabcolsep=0.05cm
    \begin{tabular}{lcccccc}
        \toprule
        Model & \multicolumn{2}{c}{GSO \cite{downs2022google}} & \multicolumn{2}{c}{Poly Haven \cite{polyhavenPolyHaven}} & \multicolumn{2}{c}{Objaverse \cite{objaverse}} \\
        \cmidrule(lr){2-3} \cmidrule(lr){4-5} \cmidrule(lr){6-7}
          & Size (MB) & Comp. & Size (MB) & Comp. & Size (MB) & Comp. \\
        \midrule
        Mesh        & 12.4 $\pm$ 2.0 & ~ & 15.2 $\pm$ 4.9 & ~ & 67.3 $\pm$ 60.6 & ~ \\
        NeRF \cite{mildenhall2020nerf}        & 5.0            & 2.5x & 5.0            & 3.0x & 5.0             & 13.5x \\
        I-NGP \cite{instantngp} & 26.4           & 0.5x & 26.4           & 0.6x & 26.4            & 2.5x \\
        Ours        & 4.4 $\pm$ 0.7  & \textbf{2.8x} & 4.9 $\pm$ 0.9  & \textbf{3.2x} & 4.7 $\pm$ 0.2   & \textbf{14.4x} \\
        \bottomrule
    \end{tabular}
    \label{tab:compression}
\end{table}

\subsubsection{Surface Normal Estimation} \label{sec:normals}
In this section, the accuracy of surface normals predicted by our proposed representation is assessed. A random sampling of 1 million points on the surface of each mesh was conducted to compare the ground-truth mesh normals with the normals predicted by the representation. As shown by Table.~\ref{tab:normals_results}, our method achieves a low root mean squared angular error (RMSAE) across the three datasets, supporting the accuracy of the predicted normals using a SIREN network with positional encoding.  While surface normal estimation is possible in NeRF-based methods by computing gradients of the volume density field, these estimates can be noisy and unreliable \cite{verbin2022ref}. Given access to ground-truth geometry, we explicitly train a module to predict surface normals from the mesh, instead of relying on gradient-based estimation.

\begin{table}[h]
    \centering
    \caption{Predicted Surface Normals RMSAE Evaluation ($\downarrow$)}
    \begin{tabular}{lc}
        \toprule
        Dataset & RMSAE (deg) \\
        \midrule
        GSO \cite{downs2022google}  & 1.99 $\pm$ 1.16  \\
        Poly Haven \cite {polyhavenPolyHaven}  & 2.05 $\pm$ 1.86\\
        Objaverse \cite{objaverse}  & 4.59 $\pm$ 1.84\\
        \bottomrule
    \end{tabular}
    \label{tab:normals_results}
\end{table}


\begin{figure}
\centering
\includegraphics[width=\linewidth]{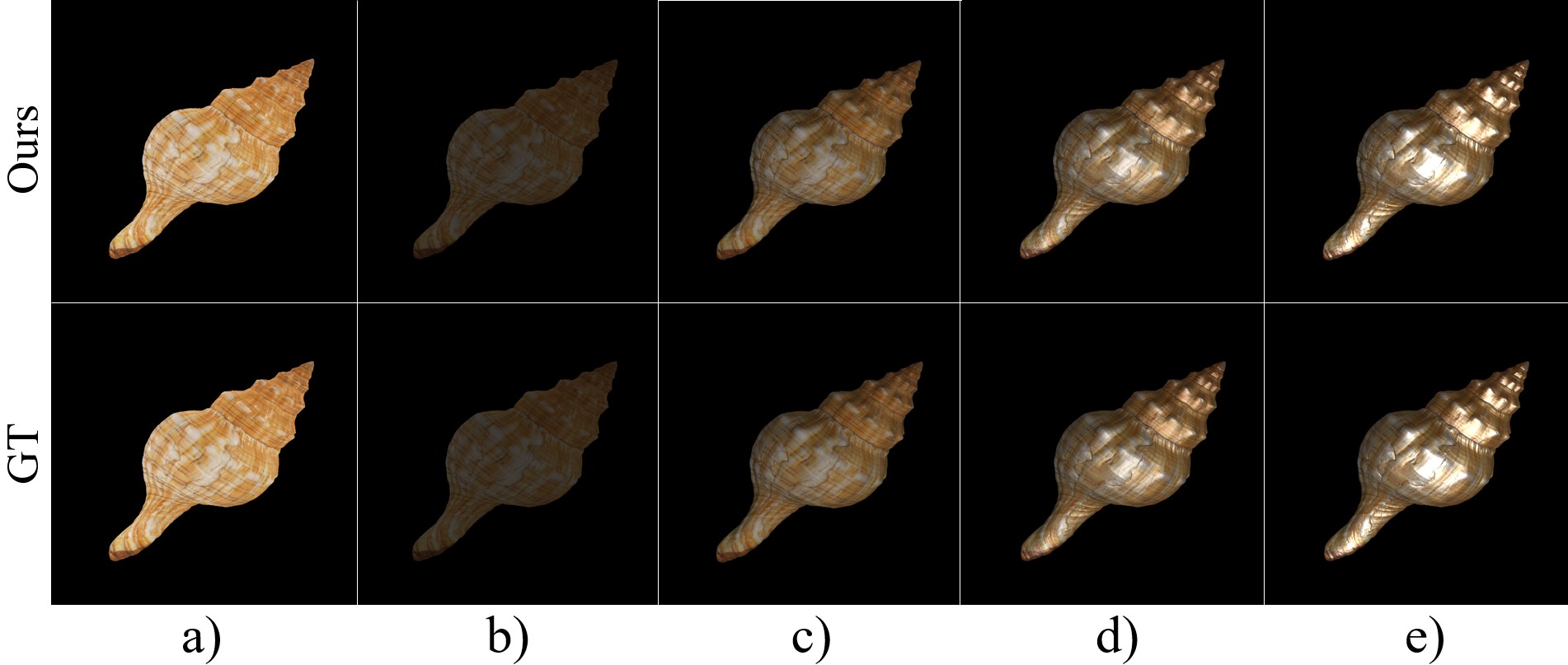} %

\caption{Relighting comparison on the Objaverse horse-conch scene \cite{objaverse}. Since our representation predicts surface normals, it supports relighting via a Blinn-Phong illumination model. \textit{Top row}: renders from our representation. \textit{Bottom row}: ground-truth mesh rendered under identical lighting. Columns show progressively composed lighting terms: (a) original scene without altered lighting, (b) ambient only, (c) ambient~+~diffuse, (d) ambient~+~diffuse~+~specular at low intensity, and (e) the same with high specular intensity.}

\label{fig:relighting} 
\end{figure}
%
%
\subsubsection{Relighting}
\label{sec:relighting}
The predicted point normals are applied to relighting mesh scenes using the Blinn-Phong model as demonstrated in Fig.~\ref{fig:relighting}. Various lighting parameters, including ambient, diffuse, and specular components were adjusted. The results show that the proposed representation effectively renders realistic lighting effects compared to the ground truth, reinforcing its potential as a viable alternative to standard textured meshes.

\subsubsection{Surface Recovery} \label{sec:extraction}
Building a robotics simulation environment requires accurate geometry for collision interactions. NeRFifyMesh preserves fine geometric details by explicitly optimizing opacity to model object surfaces. The surface was identified by sampling high-gradient regions with an alpha threshold of 0.95. These surface points and corresponding predicted normals were utilized for Poisson surface reconstruction to generate high-quality meshes post-training.

We compare our surface reconstruction results on four objects to NeRF, Instant-NGP, and NeuS. For NeRF and Instant-NGP, Marching Cubes was utilized for mesh extraction at a grid resolution of $512^3$. Extracted meshes were compared based on the Chamfer Distance metric \cite{chamferDistance} as shown in Table \ref{tab:chamfer distance}. Our representation outperforms NeRF and Instant-NGP  on all four objects, and outperforms NeuS on 3 of of the 4 objects.

In Fig.~\ref{fig:ours_vs_NerfToMesh}, NeRF and Instant-NGP produce noisy reconstructions with floating artifacts since they are designed for novel-view synthesis instead of surface reconstruction \cite{wang2023neus}. Additionally, for the Conch object, the interior features of the shell are not well reconstructed by NeRF, Instant-NGP, and NeuS due to limited visibility in the camera views. This highlights an advantage of our representation because our geometry-based sampling captures these interior features.

\begin{figure}[h]
\centering
\includegraphics[width=.47\textwidth]{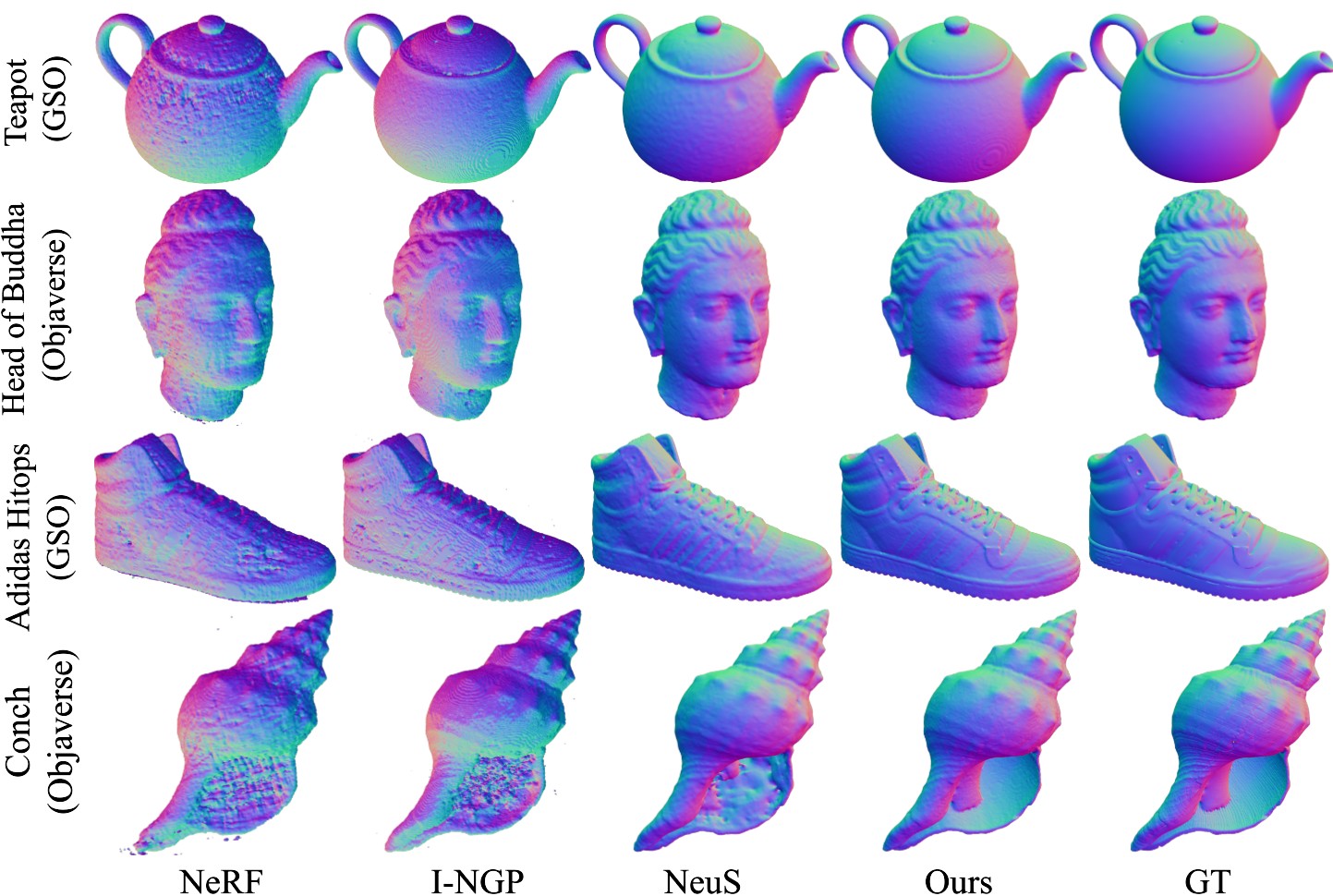} %
\caption{Surface reconstruction qualitative comparison against NeRF  \cite{mildenhall2020nerf}, Instant-NGP \cite{instantngp}, and NeuS \cite{wang2023neus}.}
\label{fig:ours_vs_NerfToMesh} 
\end{figure}

\begin{table}[h]
    \centering
    \caption{Surface Reconstruction Chamfer Distance ($\downarrow$) Comparison (units: \(10^{-3}\))}
    \begin{tabular}{lcccc}
        \toprule
        Model & NeRF \cite{mildenhall2020nerf} & I-NGP \cite{instantngp} & NeuS \cite{wang2023neus} & Ours \\
        \midrule
        Porcelain Teapot \cite{downs2022google} & 1.82 & 1.13 & 0.70 & \textbf{0.57} \\
        Head of Buddha \cite{objaverse} & 2.69 & 1.93 & 1.54 & \textbf{1.36}\\
        Adidas Hitops \cite{downs2022google} & 0.80 & 0.63 & \textbf{0.44} & 0.47 \\
        Horse-Conch \cite{objaverse} & 2.27 & 1.88 & 1.22 & \textbf{0.71} \\
        \bottomrule
    \end{tabular}
    \label{tab:chamfer distance}
\end{table}

\subsubsection{Scene Composition} 
\label{sec:composition}
\begin{figure}[h]
\centering
\includegraphics[width=\linewidth]{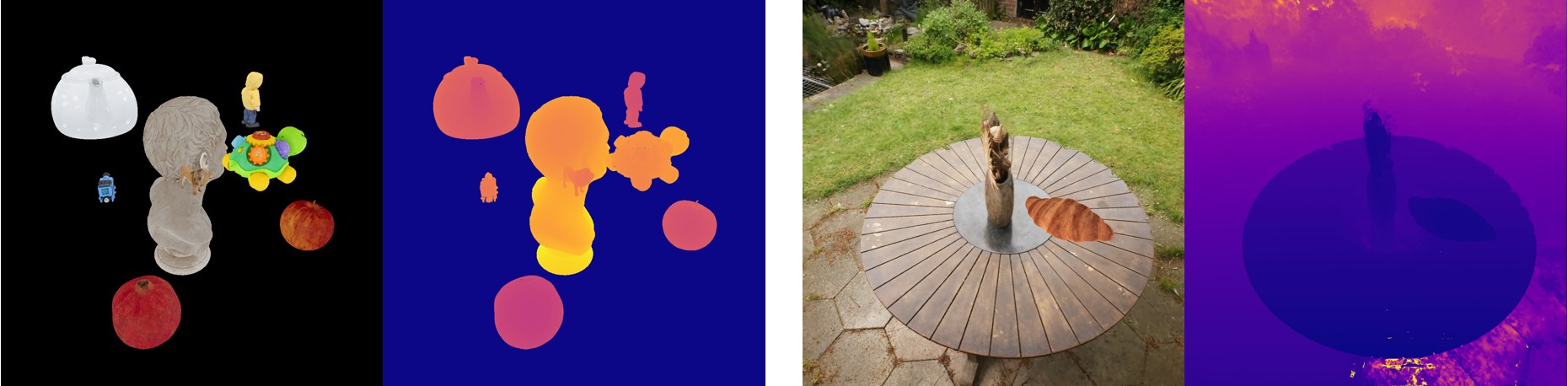} %
\caption{Scene composition results, RGB and Depth renderings. (left) A scene formed from 8 NeRFifyMesh models of objects from GSO \cite{downs2022google} and Polyhaven \cite{polyhavenPolyHaven}. (right) Intercomposition of a NeRF Garden scene (Mip-NeRF 360 dataset) \cite{barron2022mip} with a NeRFifyMesh model of a croissant from the Poly Haven dataset \cite{polyhavenPolyHaven}.
}\label{fig:composition}
\end{figure}

NeRFifyMesh is able to convert mesh-based geometry into NeRF representations, which can then be composed to form larger scenes. For example, individual NeRFifyMesh models can be transformed and placed in a global coordinate system to build a unified scene (Fig.~\ref{fig:composition}a), or they can be integrated alongside existing NeRF models (Fig.~\ref{fig:composition}b).

Each NeRFifyMesh model is normalized to fit within a unit cube and defined in a local coordinate frame. To render a global scene from these models, rays are first cast from a virtual camera in the global frame, then transformed into the local coordinate frame of each NeRFifyMesh model. Let the ray in the world frame be 
$\mathbf{r}_\text{w}(t) =\mathbf{o}_\text{w}+ t \vec{d}_{\text{w}} $, 
where \( \mathbf{o}_\text{w} \) is the ray origin and \( \vec{d}_\text{w} \) is the ray direction. To evaluate this ray in the local frame of model \( i \), the inverse transformation, 
$\mathbf{r}_i(t) = \mathbf{T}_{\text{w}i}^{-1} \mathbf{r}_\text{w}(t)$, 
is applied where \( \mathbf{T}_{\text{w}i} \in SE(3) \) is the pose of model \( i \) in the world frame.

Each local NeRF model returns RGB and depth values for the transformed rays. These per-model renderings are then composited in the 2D image plane using depth-based ordering to resolve occlusions. Specifically, for each pixel, the RGB outputs from different models are blended from back to front using the corresponding depth values. This pipeline also supports augmenting standard NeRF scenes with NeRFifyMesh models.




\subsection{Simulation Experiments}
Following the approach of NeRF2Real \cite{nerf2real}, we create dynamic simulations from our models by leveraging a physics engine (PyBullet \cite{coumans2021}). The engine is initialized with extracted meshes following the approach detailed in Sec.~\ref{sec:extraction} to form the collision representation. In addition, inertial properties, contact parameters (friction, restitution), velocities, and starting poses are provided to define the state of each object. For each rendered simulation frame, object poses from the physics engine are utilized to perform volume rendering and the rendered objects are composited to form the global scene (Sec.~\ref{sec:composition}).

In order to characterize the advantages of our representation, a collision simulation composed of two objects situated on a ground plane is constructed. We compare our results against a simulation created using Instant-NGP, where collision geometry is extracted with Marching Cubes. The ground-truth simulation uses the original mesh for collision geometry and mesh-based rendering for visualization. 

For quantitative evaluation, each rendered simulation frame is compared to the ground truth using PSNR. Trajectory accuracy is measured using mean Euclidean position error and mean rotation error, defined as the angular distance between quaternions.

Fig. ~\ref{fig:collision_sim} demonstrates visually that the simulation made with our models is a closer match to the ground truth, primarily due to improved collision geometry. This is further reinforced by the quantitative results in Table \ref{tab:collision_results}, where our representation achieves higher PSNR and lower trajectory errors (position, rotation) for both objects compared to Instant-NGP. Therefore, the improved geometry accuracy of our representation from directly sampling the mesh makes it better suited to robotics simulations. 

Additionally, simulations can be formed from static NeRF scenes and our models. Fig. ~\ref{fig:combined_sim}, shows a simulation of dropping a football (Poly Haven dataset) from a height of 0.78m onto a table within the NeRF Garden scene \cite{barron2022mip}. The static NeRF scene collision mesh was extracted using Marching Cubes and processed to remove floating artifacts. This experiment shows how our representation can facilitate combined simulations of static NeRF scenes with dynamic objects formed from meshes for robotics applications. 

\begin{figure}
\includegraphics[width=\linewidth]{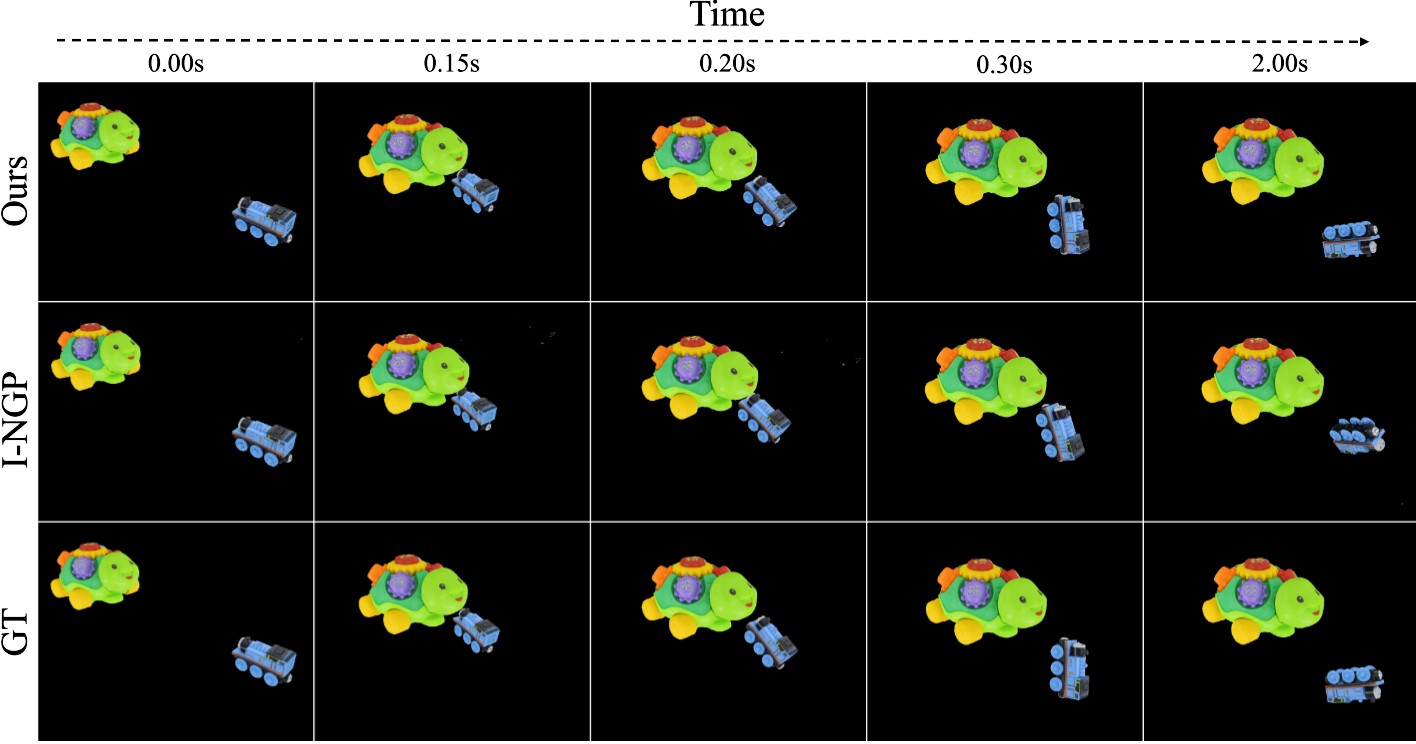} %
\caption{Visual comparison of collision simulations between the VTech Roll and Learn Turtle and Thomas the Train from GSO \cite{downs2022google}. Both objects are initialized with a velocity of 1.4 m/s on a ground plane with a friction coefficient ${\mu}_k$ = 0.8. Simulations are run for 2 seconds at 60 FPS with a physics step of 200 Hz. We compare results from our method to a simulation formed using Instant-NGP \cite{instantngp}.} 
\label{fig:collision_sim} 
\end{figure}

\begin{figure}
\includegraphics[width=\linewidth]{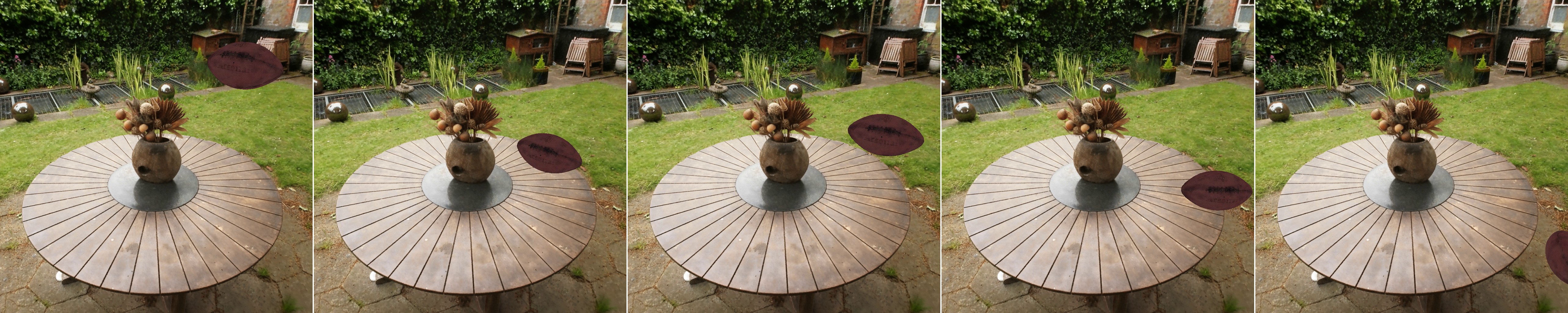} %
\caption{Collision simulation in a composite scene combining our representation of a football from Poly Haven \cite{polyhavenPolyHaven} and a static NeRF Garden scene (Mip-NeRF 360 dataset) \cite{barron2022mip}. The football was dropped from a height of 0.78m above the table in the scene.} 
\label{fig:combined_sim} 
\end{figure}

\begin{table}[ht]
\centering
\caption{Collision simulation accuracy comparison between Instant-NGP \cite{instantngp} and our method. Position error is reported as mean Euclidean error, rotation error as mean quaternion angular difference.}
\begin{tabular}{llcc}
\toprule
Object & Method & Pos. Error (m) & Rot. Error (deg) \\
\midrule
\multirow{2}{*}{Turtle \cite{downs2022google}}
    & Instant-NGP \cite{instantngp} & 1.73e-3 & 1.04 \\
    & Ours        & \textbf{7.31e-4} & \textbf{0.54} \\
\midrule
\multirow{2}{*}{Thomas \cite{downs2022google}}
    & Instant-NGP  \cite{instantngp} & 1.10e-2 & 20.34 \\
    & Ours        & \textbf{1.06e-2} & \textbf{5.99} \\
\bottomrule
\end{tabular}

\vspace{0.5em}
\small Avg. PSNR (dB): Instant-NGP = 21.35 \quad Ours = \textbf{24.60}
\label{tab:collision_results}
\end{table}

\subsection{Limitations}
A limitation of our method is its dependency on the quality of the ground-truth geometry. High face counts complicate achieving an even sampling of the mesh using face-weighted sampling. Additionally, for non-watertight meshes, the computation of the SDF becomes inaccurate, making it difficult to distinguish between the inside and outside of the mesh. These inaccuracies in SDF calculation lead to errors in the ground-truth radiance field, resulting in artifacts and noise in the scene render. Fig.~\ref{fig:limitation} illustrates the poor quality render result for the Kusunoki Masashige scan from Sketchfab \cite{takamura_koun_kusunoki_3dmodel}, which is a non-watertight mesh.
%

\begin{figure}[ht]
    \centering
    \begin{subfigure}[b]{0.4\linewidth}
        \centering
        \includegraphics[width=\linewidth, height=\textwidth]{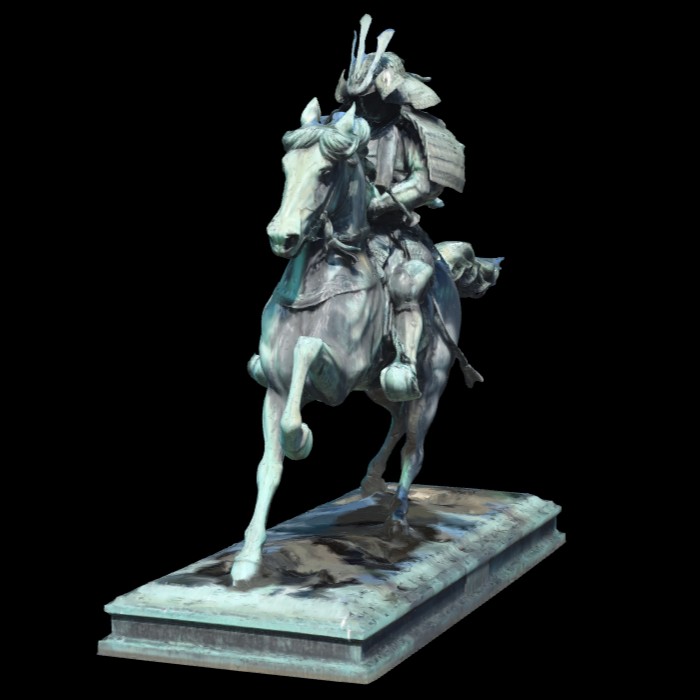}
        \caption{Ground Truth}
        \label{fig:ground_truth}
    \end{subfigure}
    \begin{subfigure}[b]{0.4\linewidth}
        \centering
        \includegraphics[width=\linewidth, height=\textwidth]{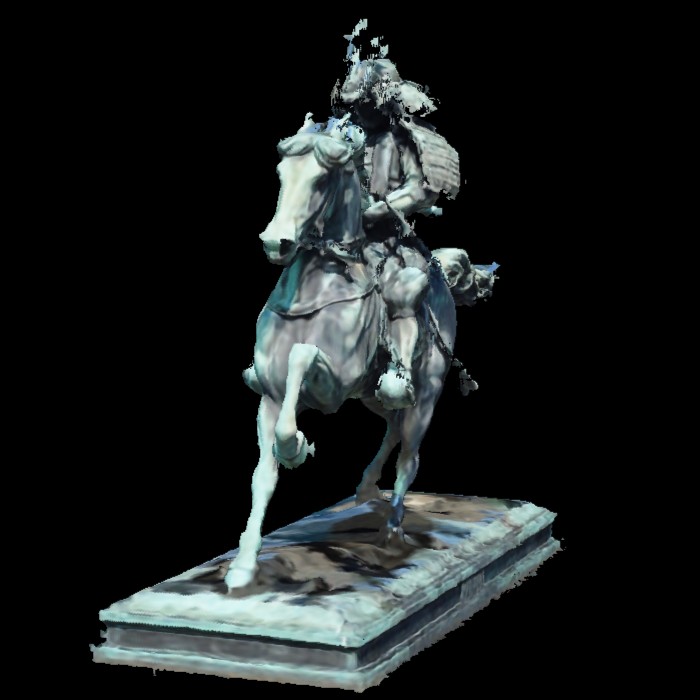}
        \caption{Ours}
        \label{fig:ours}
    \end{subfigure}
    \caption{Limitation of our method. For non-watertight meshes, inaccuracies in the SDF calculation leads to the presence of artifacts and results in poor-quality renders.}
    \label{fig:limitation}
\end{figure}

\section{Conclusion}\label{sec:con}
In this paper, we introduced NeRFifyMesh, a pipeline and architecture that directly samples a mesh to generate a ground-truth radiance field with point normals, and subsequently trains on this field to produce a NeRF. Our method achieved impressive results in mesh scene rendering compared to state-of-the-art baselines. Additionally, NeRFifyMesh supports composing multiple models into unified scenes, enabling applications in robotics simulations where constructing complex environments from reusable assets is essential.

\bibliographystyle{ieeetran}
\bibliography{egbib}

\end{document}